# Toward a Research Agenda in Adversarial Reasoning: Computational Approaches to Anticipating the Opponent's Intent and Actions

Alexander Kott[a], Michael Ownby[b]

[a]Defense Advanced Research Projects Agency, 3701 N. Fairfax Drive, Arlington VA, USA 22203
[b]Solers, Inc., 1611 N. Kent St, Arlington VA, USA 22209

## ABSTRACT

**This paper defines adversarial reasoning as computational approaches to inferring and anticipating an enemy's perceptions, intents and actions. It argues that adversarial reasoning transcends the boundaries of game theory and must also leverage such disciplines as cognitive modeling, control theory, AI planning and others. To illustrate the challenges of applying adversarial reasoning to real-world problems, the paper explores the lessons learned in the CADET -- a battle planning system that focuses on brigade-level ground operations and involves adversarial reasoning. From this example of current capabilities, the paper proceeds to describe RAID -- a DARPA program that aims to build capabilities in adversarial reasoning, and how such capabilities would address practical requirements in Defense and other application areas.**

**Keywords:** adversarial reasoning, game theory, cognitive modeling, belief and intent recognition, opponent's strategy prediction, plan recognition, deception discovery, deception planning, strategy generation

## 1. THE SCOPE OF ADVERSARIAL REASONING

This paper focuses on approaches and challenges in what is loosely encompassed by the term adversarial reasoning: computational solutions to determining the states, intents and actions of one's adversary, in an environment where one strives to effectively counter the adversary's actions.

The subtopics within this subject include belief and intent recognition, opponent's strategy prediction, plan recognition, deception discovery, deception planning, and strategy generation. From the engineering perspective, the applications of adversarial reasoning cover a broad range of practical problems: military planning and command, military and foreign intelligence, anti-terrorism and domestic security, law enforcement, information security, recreational strategy games, simulation and training systems, applied robotics, etc.

To make the term adversarial reasoning more concrete, consider the domain where it has been applied particularly extensively, the domain of military operations. In military command and control, the challenge of automating the reasoning about the intents, plans and actions of the adversary would involve the development of computational means to reason about the future enemy actions in a way that combines: the enemy's intelligent plans to achieve his objectives by effective use of his strengths and opportunities; the enemy's perception of friendly strengths, weaknesses and intents; the enemy's tactics, doctrine, training, moral, cultural and other biases and preferences; the impact of terrain, environment (including noncombatant population), weather, time and space available; the influence of personnel attrition, ammunition and other consumable supplies, logistics, communications, sensors and other elements of a military operation; and the complex interplay and mutual dependency of friendly and enemy actions, reactions and counteractions that unfold during the execution of the operation. Adversarial reasoning is the process of making inferences over the totality of the above factors.

Although many of the problems inherent in adversarial reasoning have been traditionally seen as belonging to the field of game theory, we argue here that practical adversarial reasoning calls for a broader range of disciplines: artificial intelligence planning, cognitive modeling, game theory, robust control theory, machine learning. An effective approach to the problems of adversarial reasoning must combine contributions from disciplines that unfortunately rarely come together. One of our objectives here is to demonstrate the important close relations between ideas coming from such diverse areas.

Three themes are particularly salient in adversarial reasoning. Faced with an intelligent adversary, a decision maker, whether human or computational, often must begin by using the available information in order to identify the intent, the nature and the probable plans of the adversary. Hence the first key theme of adversarial reasoning – opponent's intent and plan recognition. However, a capable adversary is likely to conceal his plans and to introduce crucial deceptions. Therefore, the second theme – deception discovery – focuses on detection of concealments and deceptions. Having made progress in identifying an adversary's intent and guarding himself against possible deceptions, the decision maker has to formulate his own plan of actions that takes into account potential counteractions of the adversary - and this is the third theme, strategy formulation.

## 2. PRACTICAL REQUIREMENTS FOR ADVERSARIAL REASONING

There are several reasons why research on adversarial reasoning is of practical interest at this time. The post-9/11 security posture of the United States fuels investments and growing interest in innovative computational techniques suitable for practical applications in military intelligence, military robotics, counter-terrorism, law enforcement and information security. For example, the Defense Advanced Research Projects Agency (DARPA) recently initiated a program in adversarial reasoning (called RAID [4]).

On the other hand, recent years have seen a dramatic rise in the capabilities of techniques relevant to adversarial reasoning, making potential solutions for the first time relevant to problems of a practical scale and complexity. The 1950's and 1960's saw critical developments in the understanding of Game Theory, a key element of adversarial reasoning. Game problems are tremendously more complex than those for systems without antagonistic inputs. Until recently, game formulations of practical problems, with the attendant level of detail and scale, resulted in a level of complexity that could not be satisfactorily handled. Today, however, there are claims of computational techniques that offer the promise of robustness and scalability appropriate for practical applications. Furthermore, there has been a dramatic rise in the maturity of technical approaches that address the cognitive aspects of adversarial reasoning, particularly the means to model how an adversary perceives a situation, reflects on what the opponent might perceive and do, and decides on a course of action.

The applied communities interested in adversarial reasoning certainly include military planners and analysts as well as the intelligence community. Those who develop applications and processes related to anti-terrorism and domestic security and law enforcement would share similar interests. Other, less obvious communities of practitioners include those concerned with financial fraud detection and information security. They also benefit from a better understanding of what and how the opponent thinks while preparing and executing a financial fraud or an intrusion into an information system. Developers of military simulation and training systems, as well as developers of commercial entertainment games, are always striving for a more realistic and intelligent opposing force within their respective systems. To a significant extent, they benefit from advances in adversarial reasoning. Finally, an even less obvious, but very relevant, area of practical applications is military robotics. In order to survive and be effective in a hostile environment, a robot (e.g., a highly autonomous unmanned aerial vehicle) must reason about the likely actions of its adversaries.

## 3. THE CADET - AN EXPLORATION IN ADVERSARIAL REASONING

One way to identify the challenges of adversarial reasoning is to look at an example of a system that requires and exhibits some elements of such reasoning. Here, for this purpose, we use the example of the Course of Action Development and Evaluation Tool (CADET), a system for semi-automated planning of US Army ground operations. Although the CADET has been discussed elsewhere in greater detail [6, 10-12], in this paper we particularly focus on those elements of the CADET that highlight the challenges of adversarial reasoning.

The CADET is a tool for automatically (or with human guidance) producing the detailed tasks required to translate a basic concept into a fully formed, actionable plan, which is a key step in the military's standard decision making process. This step involves taking the course of action for the friendly forces, developed in a previous step and initially expressed as a high-level concept, and expanding it into the hundreds of supporting tasks required to accomplish the intended objective. Crucially, it also involves estimating the reactions of the enemy, and planning appropriate counteractions. Thus, it requires adversarial reasoning.

The input for this effort usually comes from the unit Commander in the form of two doctrinally defined products: a Course of Action (COA) sketch (e.g., Fig. 1) and a Course of Action statement – a high-level textual specification of the operation. In effect, such a sketch and statement comprise a set of high-level actions, goals, and sequencing, referring largely to movements and objectives of the friendly forces, e.g., "Task Force Arrow attacks along axis Bull to complete the destruction of the 2nd Red Battalion."

The outcome of the process is usually recorded in a synchronization matrix [8], a type of Gantt chart. Time periods constitute the columns. Functional classes of actions, such as maneuver, logistics, military intelligence, etc., are the rows (see Fig. 2). The content of this plan, recorded largely in the matrix cells, includes the tasks and actions of the multiple subunits and assets of the friendly force; their objectives and manner of execution, expected timing, dependencies and synchronization; routes and locations; availability of supplies, combat losses, enemy situation and actions.

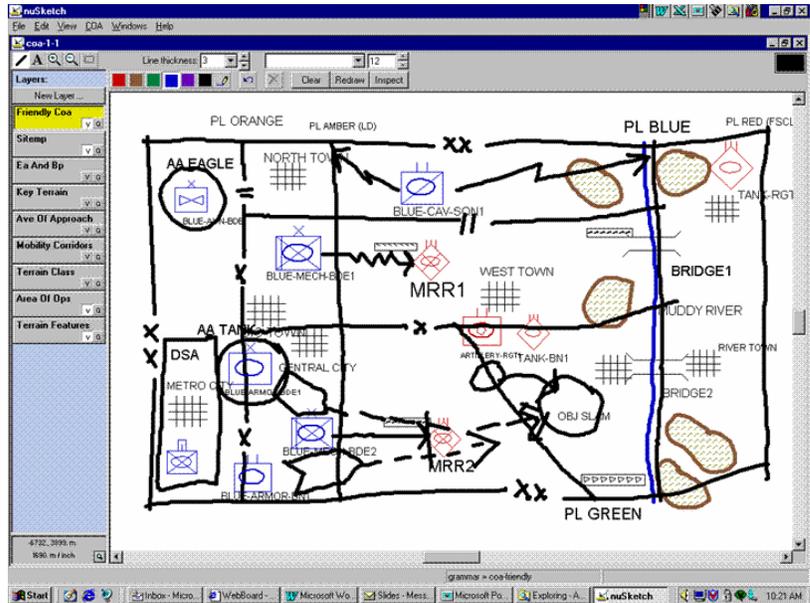

Figure 1: An example of a (partial) sketch of a course of action.

Figure 2: An example synchronization matrix (partial) produced in a planning and wargaming process, starting with the COA sketch of Fig.1 and COA statement. Such products are usually drawn by hand on a preprinted template of a synchronization matrix. More recently, these are are commonly produced with a personal computer, using conventional programs for office presentation graphics and spreadsheets.

The manual process of generating this product is complex, error-prone and time-consuming. The CADET assists military planners in this process by rapidly translating an initial, high-level COA into a detailed battle plan and wargaming the plan to determine if it is feasible. Working with the planner in a series of user/computer interactions, the system details resources, schedules, elaborates, and analyzes the COA.

In brief, the human planner defines the high-level COA via a user interface that enables him to enter the information comparable to the conventional COA sketch and statement (e.g., Fig. 1), which the COA-entry interface then transforms into an input to the CADET proper, a collection of formal assertions and/or objects, including typically on the order of 2-20 high-level tasks. This definition of the COA is transferred to the CADET, which proceeds to expand this high-level specification into a detailed plan/schedule of the operation.

Within this expansion process, the CADET decomposes friendly tasks into more detailed actions; determines the necessary supporting relations, allocates / schedules tasks to friendly assets; takes into account dependencies between tasks and availability of assets; estimates enemy actions and reactions; devises friendly counter-actions; and estimates paths of movements, timing requirements, force attrition and supply consumption. The resulting detailed, scheduled and wargamed plan often consists of up to 500 detailed actions with a wealth of supporting detail.

Having completed this process (largely automatically, in about 20 seconds on a typical laptop computer), the CADET displays the results to the user as a synchronization matrix and/or as animated movements on the map-based interface. The user then reviews the results and may either change the original specification of the COA or directly edit the detailed plan.

The technical core of the CADET is an algorithm [10] for tightly interleaved incremental planning, routing, time estimating, scheduling (partly similar to [11]), estimates of attrition [12] and consumption, and adversarial reaction estimation. This interleaving approach descends conceptually from a tree solver [9] where similar interleaving is applied to a design domain. Here we will focus on the particular technique within the overall algorithm that focuses specifically on elements of adversarial reasoning.

The CADET accounts for adversarial activity in several ways. First, it allows the commander and staff to specify the likely actions of the enemy. The automated planning then proceeds, taking into account, in parallel, both the friendly and enemy actions. Further, the CADET automatically infers (using its knowledge base and using the same expansion technique used for hierarchical task network planning) possible reactions and counteractions, and provides for resources and timing necessary to incorporate them into the overall plan. We adopted the Action-Reaction-Counteraction (ARC) heuristic technique used in the traditional COA analysis phase of the Military Decision Making Process [8]. In the Action-Reaction-Counteraction (ARC) approach, an action possible by either friendly or enemy warrants examination for potential reactions. This is followed with further analysis to determine if there exists a counter-action that can be used to minimize the impact of the reaction or negate its effects completely. The ARC technique was augmented with parallel planning for both friendly and enemy forces.

Consider the example of the activity called "forward passage of lines," in which a unit of force passes through the lines of defense manned by a friendly unit and then engages the enemy. When performing this activity, both the unit being passed, and the passing unit, are susceptible to enemy artillery fire. Therefore, the CADET's knowledge base includes a method that, in the process of expanding this activity, postulates that if the enemy has suitable forces, it will react by attacking by fire the passing unit. The method searches for enemy artillery units within the range from the location of the passage of lines, and creates the (hypothetic) enemy reaction activity "artillery fire" performed by the available enemy unit. This in turn triggers the generation of counteraction activities.

ARC does not involve an explicit search, in the sense that it does not explicitly explore multiple alternative moves at each decision point, and it does not involve backtracking, except for the user-driven backtracking. Instead, the process proceeds in a linear, depth-first fashion: for every newly generated action, the ARC method of the activity (if one is specified within the KB) produces an activity (or activities) representing the enemy reaction; the reaction activity in turn triggers a similar generation of counteraction activities.

The ARC process does not look for an optimal solution and does not guarantee one. Rather, the intent is to produce a solution that is consistent with the user's expectations and doctrinal training and is produced much faster and more accurately than in the manual process. Although ARC does not guarantee optimality, it produces solutions of the quality that experts find comparable to those of human experts. The rules that generate reactions and counteractions embody expert knowledge. The probability of generating a grossly suboptimal solution is minimized because the rules that generate a reaction do implicitly account for probable counteractions.

To evaluate the products of the CADET, we performed several series of Turing-test-like experiments [6]. In one of the experimental series, qualified human judges were asked to review two sets of battle plans: one set was generated by US Army officers during training exercises and another set was generated (for the same battle scenarios) by the CADET. The plans were disguised in a way that prevented the judges from knowing which were generated by humans and which were generated by the CADET. The objective was a "blind test" of whether there is a detectable difference between the human-generated and the CADET-generated plans. The experiment involved five different scenarios of Brigade-sized offensive operations, nine judges (all Army or Marine field grade officers - Colonels, Lieutenant Colonels or Majors - mostly active duty), four types of planning products for each scenario, and three individual grades that the judges were asked to assign to the products.

Each judge was asked to review a plan (presented as a synchronization matrix) and to provide 3 grades: one grade that characterizes the correctness and feasibility of the plan as reflected in the synchronization matrix, on the scale of 1-10; one grade that characterizes the completeness and thoroughness of the plan as reflected in the synchronization matrix, on the scale of 1-10; and one qualitative grade that compares the plan with typical products they see in today's Army practice (ranging from "much worse" to "much better"). Instructions to the judges were worded in a way that associated the grade of 5 with the "typical quality found in today's practice."

Overall, the results demonstrated that the CADET performed on par with the human staff - the difference between the CADET's and a human's performance was statistically insignificant. Taking the mean of the grades for all five scenarios, the CADET earned 4.2, and humans earned 4.4, with standard deviation of about 2.0, a very insignificant difference. However, it typically takes a human staff about 16 person-hours to produce a plan, while the CADET assisted user spends only about 0.4 person-hours per plan. The conclusion: the CADET helped produce complex planning products dramatically faster (almost two orders of magnitude faster) yet without loss of quality, as compared to the conventional, manual process.

From the perspective of adversarial reasoning, the judges never felt that the CADET was insufficiently mindful of the adversary. Overall, the ARC technique produced battle plans that look to human reviewers rather sophisticated, insightful and proactive: they seemed to anticipate and parry enemy actions multiple "moves" ahead, even though in fact the ARC technique does not involve any look-ahead reasoning.

So much for the strong points of the CADET, now let us take a look at its shortcomings. First, the CADET has no mechanism for explicit look ahead, for wargaming or game-solving in any sense. Although the CADET-generated plans seem to anticipate enemy actions and do detailed preparations for such actions far in advance, it is largely an illusion. The CADET plans backwards from the key events pre-defined in the human-generated high-level course of action. Unlike humans, the CADET does not attempt to invent (even in a limited sense) the strategy of the battle. It merely fills in the details (albeit important and complicated) into the outline of an adversarial encounter envisioned by the human. While acceptable in some applications, this shortcoming may not be in many others. Further, much of warfare is based on deception and concealment. Nothing in the CADET explicitly reasons about such issues.

The CADET also has no means to take into account the emotional and cognitive aspects of the battle. Real human warriors, at all levels of responsibility, have beliefs, emotions, desires, biases, preferences, etc., that contribute much to their plans and actions. These aspects are complex, and they change dynamically as the battle unfolds. The CADET does not reason on such factors. It also does not take into account the inevitable errors and cognitive limitations of the humans in real-world warfare.

Finally, the CADET does not take into consideration a very important factor: the impact of decision-making processes and organizations on the enemy (and friendly) actions. There are complex and influential dynamics in command decision-making, in communications, in the propagation of uncertainty, errors, confusion, trust and fears through the formal and informal network of leaders of multiple units and echelons. The problem becomes even more complex when broader, non-military societal concerns and phenomena – political, financial, ideological, etc. – must be taken into account [13]. All of this is outside the CADET's scope.

The CADET, of course, is not alone in suffering from such shortcomings. It merely indicates the state of the art in practical adversarial reasoning.

## 4. RAID – THE NEAR-TERM FUTURE OF ADVERSARIAL REASONING

If the best of today's capabilities is exemplified by tools like the CADET, the near-future may be defined by the ongoing DARPA program called RAID. Started in the fall of 2004, the Real-time Adversarial Intelligence and Decision-making (RAID) program focuses on the challenge of anticipating enemy actions in a military operation. In a number of recent publications, US military leaders call for the development of techniques and tools to address this critical challenge.

The US Air Force community uses the term predictive battlespace awareness [1-2] while a related term, predictive analysis, is beginning to be used in the US Army community [3]. Both refer to future techniques and technologies that would help the commander and staff to characterize and predict likely enemy courses of action, to relate the history of the enemy's performance to its current and future actions, and to associate these predictions with opportunities for friendly actions and effects. Both communities have pointed out the lack of technologies, techniques and tools to support predictive analysis and predictive battlespace awareness.

The RAID program aims to produce key technologies for tools capable of in-execution predictive analysis of an enemy's probable actions. A particular focus of the program will be tactical urban operations against irregular combatants (Fig. 3) – an especially challenging and operationally relevant domain.

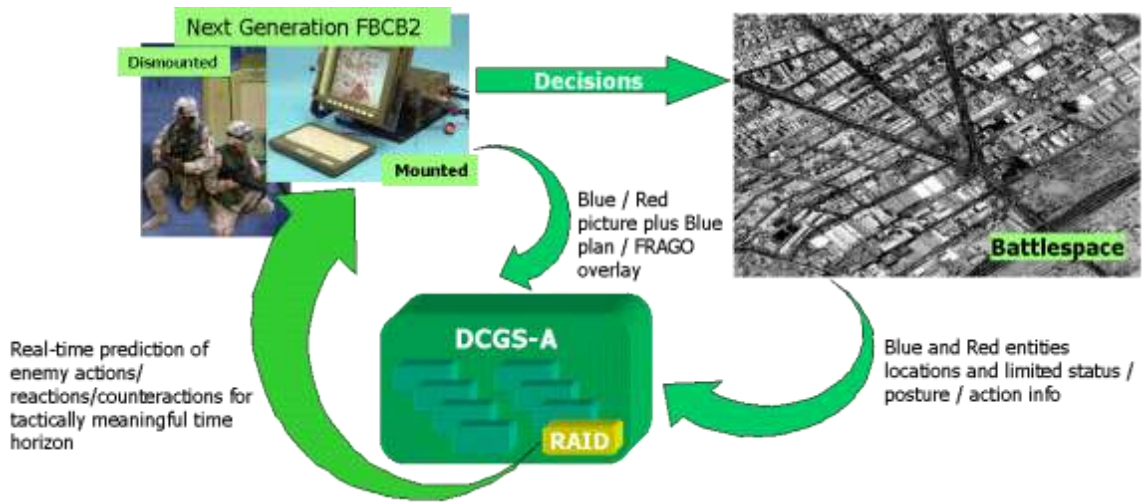

Figure 3: One possible application of RAID is a component within the future military intelligence systems. RAID would provide predictive running estimates of the enemy situation.

The program intends to leverage novel approximate game-theoretic, deception-sensitive algorithms and cognitive modeling to provide real-time enemy estimates to a tactical commander. In doing so, the RAID program will address two critical technical challenges: (a) Adversarial Reasoning: the ability to continuously identify and update predictions of likely enemy actions; and (b) Deception Reasoning: the ability to continuously detect likely deceptions in the available battlefield information (Fig. 4).

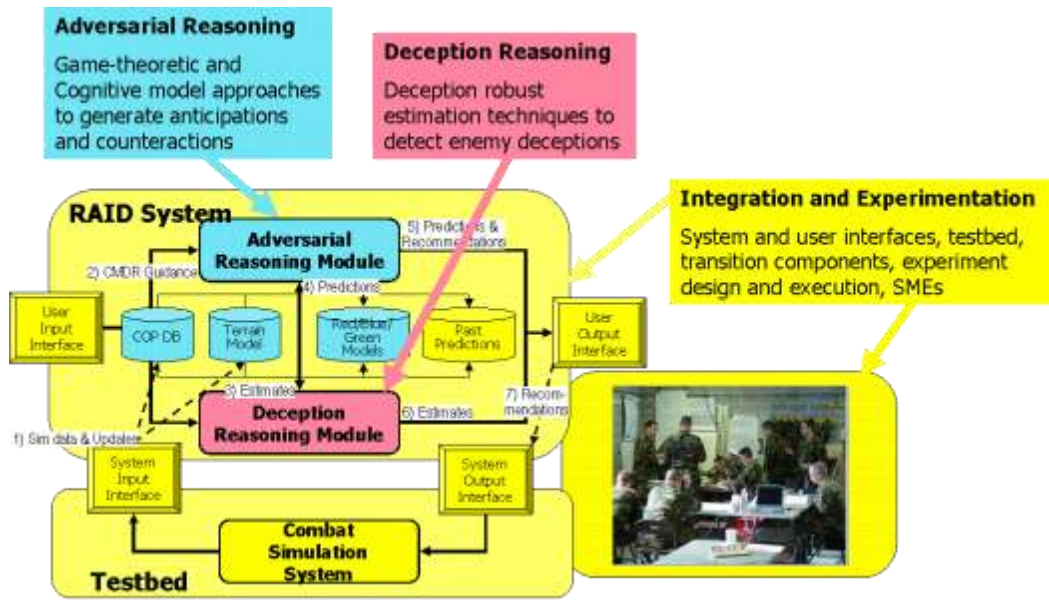

Figure 4: Key components and technical approaches of the RAID program.

Realistic experimentation and evaluation will drive the development process using human-in-the-loop, simulation-based wargames to compare humans and RAID in their adversarial reasoning capabilities.

The overarching technical challenges of this problem include some of those we explored in the CADET, but also include others. There is a tight interdependence, coupling of blue and red actions. Blue knowledge of red assets and actions is inevitably limited. Observations as well as interpretations of the observations are subject to a significant degree of errors and latency. In addition to partial, delayed and often erroneous observations, the knowledge of the battlefield is limited by a purposeful, continuous, aggressive, intelligent concealment and deception. Due to cultural, doctrinal, and psychological effects, it is not enough to only consider the most dangerous (theoretically optimal to the enemy) course of action. The actual, most likely course of action can be affected by a broad range of "human factors" and be significantly different from the theoretically most advantageous one. Complex urban terrain offers a high density as well as a fragmentation of threats and opportunities for forces [7]. Further, the terrain itself is dynamic because it is continually modified by human actions (barricades in the stress, holes in the walls, etc.).The presence of non-combatants on the battlefield must be explicitly considered and collateral damage minimized. Fire and maneuver of forces are not the only actions that must be carefully considered. Intelligence gathering, communications, and logistics (including casualty evacuation) are tightly coupled with fire and maneuver. The scale of the computational problem is immense and yet solutions must be generated in near real-time.

The RAID program considered a number of technical approaches to overcoming the above mentioned challenges. None of them are without their difficulties. Examples include the following:

- **Game-theoretic and game-playing approaches**: devising sequences of actions for both red and blue forces in a manner that assumes both sides strive to maximize the achievement of their respective objectives. Such approaches must pay special attention to the need for solving very large scale problems in near real time, recognizing the stochastic nature of outcomes for most moves, and addressing partial observability and deception issues.

- **Adversarial planning**: forming plans for both red and blue actions that lead to the achievement of the respective desired goals while preventing the attainment of the goals of the other side; often using significant amounts of domain-specific knowledge. An important challenge in the application of such approaches is to make use of relatively few elements of domain-specific knowledge, as domain-independent as possible and easy to acquire, modify and manage.

- **Deception discovery**: analyzing the information state from a risk-sensitive perspective to determine which alternative hypotheses would benefit the enemy the most if accepted by the friendly forces; analyzing the significance of preconditions for the feasibility of alternative enemy courses of action to identify the one that are more likely to be the subject of deceptions; comparing earlier expectations with current evidence to find unexplainable deviations. Such approaches would have to find ways to deal with the complex, multi-dimensional nature of the RAID problem; to work without the benefit of relying on significant amount of the human analyst's input (if any), and cooperate with an adversarial reasoning component that may use a very different representational paradigm.

- **Pattern recognition**: identifying patterns and anomalies in spatial and temporal locations, movements and other actions of the red force that could indicate concealment, deception and future intended courses of action; often using learning techniques to build and extend the repertoire of such patterns. Among the challenges relevant to such approaches are the need for effective generalization, especially in very complex terrain; the need to align pattern analysis with the enemy's objectives and goals; and ways to prevent the red from using such pattern recognition means as an effective approach to deceiving the blue.

## 5. CONCLUSIONS

The demands of practical applications call for next generation of capabilities in adversarial reasoning. A broad, interdisciplinary field of study, adversarial reasoning includes belief and intent recognition, opponent's strategy prediction, plan recognition, deception discovery, deception planning, and strategy generation. From the engineering perspective, the applications of adversarial reasoning cover a broad range of practical problems: military planning and command, military and foreign intelligence, anti-terrorism and domestic security, law enforcement, information security, recreational strategy games, simulation and training systems, applied robotics. Recent years have seen a dramatic rise in capabilities of techniques relevant to adversarial reasoning, making potential solutions for the first time relevant to problems of practical scale and complexity. Many of the more fundamental challenges include the human (both individual and organizational) elements of adversarial reasoning: human beliefs, desires and emotions; human cognitive limitations and decision-making processes.